\documentclass[5p, times, preprint]{elsarticle} % '5p' para um layout de coluna dupla

\usepackage[utf8]{inputenc}
\usepackage[english]{babel}
\usepackage[T1]{fontenc}
\usepackage{lmodern}
\usepackage{graphicx}
\usepackage{indentfirst}
\usepackage{color}
\usepackage{microtype}
\usepackage{amsmath}
\usepackage{helvet}
\usepackage{natbib}
\usepackage{url} 
\usepackage{multirow}
\usepackage{multicol}
\makeatletter
\def\subsection{\@startsection{subsection}{2}{\z@}{18\p@ \@plus 6\p@ \@minus 3\p@}{9\p@}{\normalfont\normalsize\bfseries}}
\makeatother

\makeatletter
\def\subsubsection{\@startsection{subsubsection}{2}{\z@}{18\p@ \@plus 6\p@ \@minus 3\p@}{9\p@}{\normalfont\normalsize\bfseries}}
\makeatother

\usepackage{tikz}
\usetikzlibrary{shapes.geometric, arrows}
\tikzstyle{startstop} = [rectangle, rounded corners, minimum width=1cm, minimum height=0.8cm,text centered, draw=black, fill=red!30, font=\footnotesize]
\tikzstyle{io} = [minimum width=1cm, minimum height=0.8cm, text centered, draw=black, fill=blue!30, font=\footnotesize]
\tikzstyle{process} = [rectangle, minimum width=2cm, minimum height=0.8cm, text centered, draw=black, fill=orange!30, font=\footnotesize]
\tikzstyle{decision} = [diamond, minimum width=2cm, minimum height=0.8cm, text centered, draw=black, fill=green!30, font=\footnotesize]
\tikzstyle{arrow} = [thick,->,>=stealth]
\begin{document}

\begin{frontmatter}

\title{Multi-level Product Category Prediction through Text Classification}

% Endereço de afiliação compartilhado
\address[icmc]{Institute of Mathematics and Computer Sciences, University of São Paulo, São Carlos, Brazil}
\address[ufscar]{Department of Statistics, Federal University of São Carlos, S\~ao Paulo, Brazil}

% Primeiro autor
\author[icmc]{Wesley Ferreira Maia}
\ead{wesley.ferreira.souza@usp.br}

% Segundo autor
\author[icmc]{Angelo Carmignani}
\ead{angelo.carmignani@usp.br}

% Terceiro autor
\author[icmc]{David Luz}
\ead{davidluz@usp.br}

% Quarto autor
\author[icmc]{Gabriel Bortoli}
\ead{gbortoli@usp.br}

% Quinto autor
\author[icmc]{Lucas Maretti}
\ead{lucas.maretti@usp.br}

% Sexto autor
\author[icmc,ufscar]{Daniel Camilo Fuentes Guzman}
\ead{camiguz89@usp.br}

% Sétimo autor
\author[icmc,ufscar]{Marcos Jardel Henriques}
\ead{jardel@usp.br}

% Oitavo autor
\author[icmc]{Francisco Louzada}
\ead{louzada@icmc.usp.br}

\cortext[cor1]{Corresponding author}

\begin{abstract}
This article investigates applying advanced machine learning models, specifically LSTM and BERT, for text classification to predict multiple categories in the retail sector. The study demonstrates how applying data augmentation techniques and the focal loss function can significantly enhance accuracy in classifying products into multiple categories using a robust Brazilian retail dataset. The LSTM model, enriched with Brazilian word embedding, and BERT, known for its effectiveness in understanding complex contexts, were adapted and optimized for this specific task. The results showed that the BERT model, with an F1 Macro Score of up to $99\%$ for segments, $96\%$ for categories and subcategories and $93\%$ for name products, outperformed LSTM in more detailed categories. However, LSTM also achieved high performance, especially after applying data augmentation and focal loss techniques. These results underscore the effectiveness of NLP techniques in retail and highlight the importance of the careful selection of modelling and preprocessing strategies. This work contributes significantly to the field of NLP in retail, providing valuable insights for future research and practical applications.
\end{abstract}


\end{frontmatter}

\section{Introduction}

In recent years, text classification has emerged as a powerful tool in the retail industry, assisting companies in organizing and analyzing large volumes of unstructured data \cite{kowsari2019text}. With the exponential growth of data generation, especially in unstructured formats such as text, audio, and video, efficiently classifying this data has become crucial for success in the retail sector.

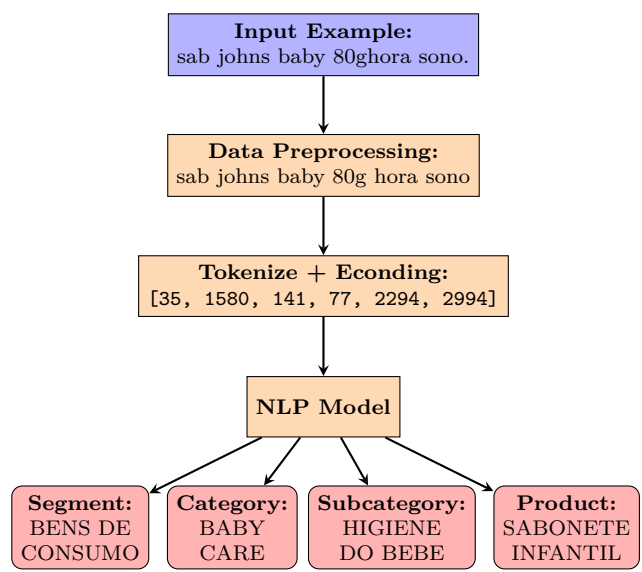
\begin{figure}[ht]
    \centering
    \begin{tikzpicture}[node distance=1.6cm]
        \node (input) [align=center, io] {\textbf{Input Example:}\\sab johns baby 80ghora sono.};
        \node (processing) [align=center, process, below of=input] {\textbf{Data Preprocessing:}\\sab johns baby 80g hora sono};
        \node (data) [align=center, process, below of=processing] {\textbf{Tokenize + Econding:}\\\texttt{[35, 1580, 141, 77, 2294, 2994]}};
        \node (bert) [process, below of=data, xshift=0 cm] {\textbf{NLP Model}};

        \node (categoria) [startstop, align=center, below of=bert, xshift=-3.2 cm] {\textbf{Segment:}\\BENS DE\\CONSUMO};
        \node (segmento) [startstop, align=center, below of=bert, xshift=-1.2cm] {\textbf{Category:}\\BABY\\CARE};
        \node (subcategoria) [startstop, align=center, below of=bert, xshift=0.9cm] {\textbf{Subcategory:}\\HIGIENE\\DO BEBE};
        \node (produto) [startstop, align=center, below of=bert, xshift=3.2cm] {\textbf{Product:}\\SABONETE\\INFANTIL};

        \draw [arrow] (input) -- (processing);
        \draw [arrow] (processing) -- (data);
        \draw [arrow] (data) -- (bert);
        
        \draw [arrow] (bert) -- (categoria);
        \draw [arrow] (bert) -- (segmento);
        \draw [arrow] (bert) -- (subcategoria);
        \draw [arrow] (bert) -- (produto);
    \end{tikzpicture}
    \caption{Example of the text classification to predict product categories in multi-categories.}
    \label{fig:general_model}
\end{figure}

Applying natural language processing (NLP) and machine learning techniques to retail data enhances the user experience on e-commerce platforms and provides valuable insights for more informed decision-making. Textual data analysis, such as customer reviews and product descriptions, enables a better understanding of consumer preferences and behaviours \cite{li2022text, huang2023sentiment, vanaja2018aspect}.

In this study, we focus on the implementation of neural networks, specifically the BERT \cite{Devlin2019BERTPO} (Bidirectional Encoder Representations from Transformers) model and LSTM \cite{10.1162/neco.1997.9.8.1735} (Long Short-Term Memory) model to address the challenge of text classification to predict product categories at multiple levels. These advanced machine learning technologies are chosen due to their proven ability to process and understand natural language, enabling more accurate and efficient classification.

In this study, we utilized the "Classificação Produtos Varejo CPG PTBR" (Retail CPG Product Classification) dataset, provided by Gilsiley Henrique Darú on Kaggle\footnote{https://www.kaggle.com/dsv/4265348}, which provides a rich and detailed dataset tailored for the retail industry. This dataset has enabled us to apply our modelling techniques to practical and relevant retail industry scenarios. The use of this dataset not only emphasizes the validity of our research but also offers valuable insights into the application of advanced technologies in real-world retail contexts.

An innovative aspect of this study was using data augmentation techniques and the focal loss function in modelling \cite{lin2017focal}. These approaches were adopted to enhance the model's effectiveness in dealing with the complexities and nuances of retail text data. The result was a remarkable performance in the evaluation metric, highlighting the potential of these techniques in improving automatic text classification in the retail sector.

This study pays special attention to multi-category text classification, a complex task that requires a detailed understanding of various product categories and subcategories. Figure \ref{fig:general_model} presents the overall model diagram used in this study, illustrating an example of how product categories are processed and classified. This model demonstrates the ability to handle the complexity and variety of retail data, offering a comprehensive solution for accurately classifying products at multiple levels.

\subsection{Structure of the Work}
The work is structured as follows: After this Introduction, the Literature Review section provides an overview of relevant approaches and technologies in NLP and text classification, emphasising the retail context. The Data and Methodology section describes the dataset, preprocessing techniques, and implemented machine learning models. In Results and Discussion, we present the obtained results and discuss their implications, headlining the effectiveness of the approaches used. Finally, the Conclusion and Future Work section summarizes the main findings and suggests directions for future research in this field.

\section{Literature Review}

Automatic text classification has become a growing interest in the NLP community, mainly due to its applicability in various domains, such as retail. This section reviews the literature, featuring significant developments and approaches adopted for text classification in retail contexts. With the advent of deep learning models, such as RNNs and the BERT model, the ability to efficiently and accurately process and classify text has advanced significantly. These models offer improvements in accuracy and language processing capability and open up new possibilities for text data analysis in the retail sector.

RNNs, especially LSTM variations, have demonstrated remarkable effectiveness in text sequence classification within NLP. These networks can capture long-range dependencies in text, which is essential for understanding the structure and context of sentences and paragraphs. This aspect is particularly relevant in retail, where accuracy in interpreting product descriptions and customer reviews can directly impact product categorization and sentiment analysis \cite{osti_6910294, doi:10.1073/pnas.79.8.2554, 10036794, 10.1162/neco.1997.9.8.1735}.

Furthermore, introducing the BERT model, which uses Transformer techniques to process words in a bidirectional context, has revolutionized text classification approaches. BERT and its derivatives have demonstrated remarkable performance in various NLP tasks, including text classification. Its ability to understand language context and nuances makes it particularly suitable for dealing with the unique challenges presented by retail texts, which often include jargon, colloquial language, and complex sentence structures \cite{Devlin2019BERTPO, 10391239, 8996183}.

In addition to neural network architectures, another focus in the literature is using data augmentation techniques and loss optimization methods, such as focal loss. These techniques are essential for improving model performance on imbalanced datasets, a common challenge in retail where specific product categories or sentiments may be more prevalent than others \cite{zhu2004personalized, gonzalez2020comparing, lin2017focal}.

Apart from theoretical approaches and neural network models, it is essential to consider the practical applications of text classification in the retail sector. Several case studies demonstrate the effective use of NLP and text classification techniques to improve inventory management, personalize the customer experience, and optimize marketing operations. For example, customer reviews and feedback analysis have been used to categorize products more effectively and identify market trends \cite{kowsari2019text, dalal2011automatic, 10036794}.

Applying NLP and machine learning techniques to retail data enhances the user experience on e-commerce platforms and provides valuable insights for more informed decision-making \cite{Chowdhary2020NLP}. Textual data analysis, such as customer reviews and product descriptions, allows a better understanding of consumer preferences and behaviours.

In addition to the mentioned advancements, another crucial aspect of text classification in the retail context is the classification of Product Categories at Multiple Levels. This approach involves organizing products into a hierarchy of categories, ranging from general to specific. For example, a product can be categorized at a higher level as "Electronics", at an intermediate level as "Televisions", and at a more specific level as "4K Televisions". This multi-level approach is particularly challenging due to the inherent complexity of hierarchical structure and the need for precision at each categorization level \cite{10.5555/3507788.3507794, Ozyegen2022}.

While advancements in text classification bring significant benefits, they also present specific challenges and limitations that are crucial to understand and address. One of the main challenges is dealing with imbalanced data, a common occurrence in retail where specific product categories or types of customer feedback are more predominant than others. This problem can lead to biased classification models favouring the majority classes at the expense of the minority ones, resulting in a distorted understanding of the market and consumer preferences \cite{kozareva2015everyone, kannan2011improving}.

Additionally, the ambiguity and variability of natural language in product descriptions and customer reviews pose additional challenges. The use of jargon, slang, and regional expressions can affect the accuracy of text classification, requiring more robust and adaptive models that can understand and interpret a wide range of linguistic expressions \cite{wei2020deep}.

Data privacy and security are also growing concerns, especially with the increasing use of customer data for text analysis. Ensuring that text classification models are developed and used ethically and responsibly is crucial, respecting consumer privacy and data regulations \cite{liu2018towards}.

\section{Data and Methodology}

In this study, we adopt a data-driven and methodological approach to evaluate the performance of LSTM and BERT neural network models in text classification, particularly for predicting various categories within the retail sector. The dataset, available on Kaggle and titled "Classificação Produtos Varejo CPG PTBR" by Gilsiley Henrique Darú, comprises descriptions of retail products. Our methodology\footnote{https://github.com/MECAI2022/short\_text\_classification} is meticulously designed, adhering to established best practices in machine learning and natural language processing. This approach is intended to ensure the accuracy of our findings and enhance their relevance and applicability to real-world scenarios in the retail industry.

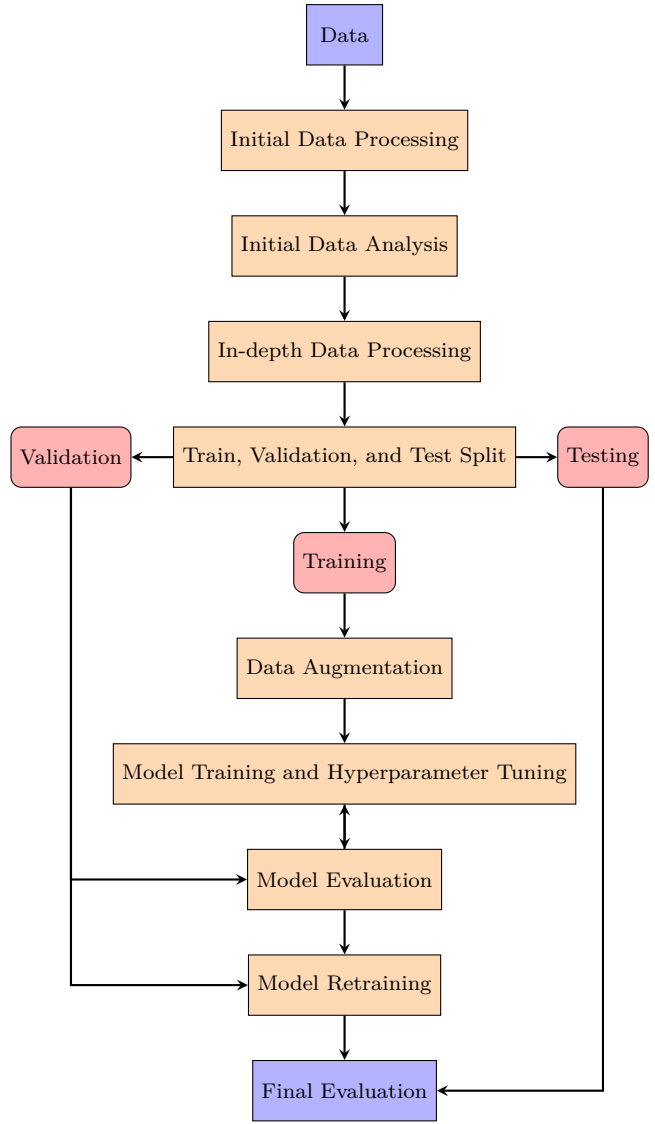
\begin{figure}[ht]
    \centering
    \begin{tikzpicture}[node distance=1.4cm]
        \node (data) [io] {Data};
        \node (Trat_init) [process, below of=data] {Initial Data Processing};
        \node (anal_init) [process, below of=Trat_init] {Initial Data Analysis};
        \node (Trat_aprof) [process, below of=anal_init] {In-depth Data Processing};
        \node (sep_treino) [process, below of=Trat_aprof] {Train, Validation, and Test Split};
        \node (train) [startstop, below of=sep_treino] {Training};
        \node (test) [startstop, right of=sep_treino,xshift=2cm] {Testing};
        \node (val) [startstop, left of=sep_treino,xshift=-2.2cm] {Validation};
        \node (aug) [process, below of=train] {Data Augmentation};
        \node (train_model) [process, below of=aug] {Model Training and Hyperparameter Tuning};
        \node (eval_model) [process, below of=train_model] {Model Evaluation};
        \node (re_train) [process, below of=eval_model] {Model Retraining};
        \node (final_eval) [io, below of=re_train] {Final Evaluation};
    
        \draw [arrow] (data) -- (Trat_init);
        \draw [arrow] (Trat_init) -- (anal_init);
        \draw [arrow] (anal_init) -- (Trat_aprof);
        \draw [arrow] (Trat_aprof) -- (sep_treino);
        
        \draw [arrow] (sep_treino) -- (train);
        \draw [arrow] (sep_treino) -- (test);
        \draw [arrow] (sep_treino) -- (val);
        
        \draw [arrow] (train) -- (aug);
        \draw [arrow] (aug) -- (train_model);
        \draw [arrow] (train_model) -- (eval_model);
        \draw [arrow] (val) |- (eval_model);
        \draw [arrow] (eval_model) --  (train_model);
        
        \draw [arrow] (eval_model) -- (re_train);
        \draw [arrow] (val) |-  (re_train);
        
        \draw [arrow] (re_train) -- (final_eval);
        \draw [arrow] (test) |-  (final_eval);
    \end{tikzpicture}
    \caption{Diagram of the ETL process used in the study of text classification to predict multiple categories in the retail sector.}
    \label{fig:etl_process}
\end{figure}

In line with the comprehensive study on data science pipelines by Biswas, Wardat, and Rajan \cite{10.1145/3510003.3510057}, our process begins with a thorough data preprocessing stage, as illustrated in the ETL diagram in Figure \ref{fig:etl_process}. This stage, essential in any data science pipeline, includes data cleaning and normalization and preparing the dataset for analysis. Exploratory Data Analysis (EDA) is then performed to understand the data's characteristics better, guiding subsequent modelling and analysis decisions.

The selection and configuration of classification models, including BiLSTM and BERT, are crucial for the study. These models are evaluated for their ability to classify short texts into different product categories, using techniques such as word embedding and cross-validation to enhance their performance.

The effectiveness of the models is exclusively assessed through the F1-score metric, providing a detailed analysis of their text classification performance. This evaluation process follows the steps described in the ETL diagram in Figure \ref{fig:etl_process}, including model training, validation, and testing, culminating in a final evaluation.

This study applies advanced NLP and machine learning methods within a structured workflow, as supported by recent research \cite{10.1145/3510003.3510057}, ensuring reliable and relevant results for the retail sector.

\subsection{Dataset}

The dataset \cite{gilsiley_henrique_darú_2022} used in our study is a comprehensive collection of 153,445 items showcasing a diverse array of retail products. It is meticulously categorized into various segments, including detailed classifications such as item name (\textit{nm\_item}), segment, category, subcategory, and product name (\textit{nm\_product}). The structure and depth of this dataset mirrors the complexity and variety inherent in the retail sector.

The data distribution is as follows: 6 different segments, 70 categories, 153 subcategories, and 715 distinct product names. This data structure offers a comprehensive view of the retail market and provides a rich foundation for text classification analyses.

It is important to note that this dataset is specific to the Brazilian market, which makes it particularly relevant for analyzing retail trends and patterns in Brazil.

\begin{table}[ht]
\centering
{\fontfamily{phv}\selectfont} % Inicia a fonte Helvetica
\caption{Structure of the Dataset}
\tiny
\begin{tabular}{p{1.2cm}|p{1.2cm}|p{1.5cm}|p{1.5cm}|p{1.3cm}}
\hline
\textbf{nm\_item} & \textbf{Segment} & \textbf{Category} & \textbf{Subcategory} & \textbf{nm\_product} \\ \hline
cueca sunga lupo g817 & FASHION E ESPORTIVO & VESTUARIO & MODA INTIMA & CUECA \\ \hline
carne suin espinhaco 1kg & BENS DE CONSUMO & ACOUGUE E PEIXARIA & ACOUGUE & CARNE SUINA \\ \hline
whisky johnn walker & BENS DE CONSUMO & BEBIDAS ALCOOLICAS & DESTILADOS & WHISKY \\ \hline
... & ... & ... & ... & ... \\ \hline
\end{tabular}
\label{tab:dataset_structure}
\end{table}

\subsection{Data Preprocessing}

Data preprocessing is crucial in preparing the dataset for subsequent analysis and machine learning techniques. In this research, the initial retail product descriptions dataset underwent a multifaceted preprocessing process to ensure data quality and consistency before applying machine learning techniques.

\paragraph{Initial Cleaning and Formatting}
Initially, the data was received in Excel format (.xlsx) and converted to a CSV file using the delimiter ';'. This conversion was necessary to facilitate data manipulation in programming environments like Python. The first step in Excel involved correcting column misalignments and checking data type consistency. Special attention was given to ensuring that text fields did not contain formatting errors, such as extra spaces, unintended special characters, and inconsistencies in capitalization.

\paragraph{Handling Missing Data and Anomalies}
After importing the data into the Python environment, an analysis was conducted to identify and address missing data. Using the Pandas library, we identified that only a tiny fraction of entries (less than 0.1\%) had missing values. Given the statistical insignificance of this data, we chose to exclude them from the dataset. Furthermore, checks were performed to identify and correct potential anomalies, such as duplicate entries or outliers, that could distort the analysis.

\subsection{Text Normalization and Standardization}

Text normalization and standardization are crucial steps in NLP data preprocessing, designed to transform raw text into a form that is homogeneous and suitable for systematic analysis, reducing its variability and complexity.

\paragraph{Conversion to Lowercase and Removal of Special Characters}
The initial step involves converting all text to lowercase and removing special characters. Let \( \mathcal{T} \) represent the text corpus, consisting of a sequence of characters \( X = x_1, x_2, \dots, x_n \). The transformation to lowercase is formalized as:

\begin{align}
\alpha(X) = \ & (x'_1, x'_2, \dots, x'_n) \nonumber \\
& \text{where} \quad x'_i = \text{lower}(x_i) \quad \forall x_i \in X
\end{align}

Here, \( \text{lower}(x_i) \) denotes the operation of converting character \( x_i \) to its lowercase equivalent \( x'_i \). The removal of special characters, represented by the function \( \beta: \mathcal{T} \rightarrow \mathcal{T} \), eliminates non-alphanumeric characters:

\begin{align}
\beta(X) = \text{clean\_chars}(X)
\end{align}

\paragraph{Extraction and Normalization of Measurement Units}
The process of identifying and normalizing measurement units involves using pattern recognition techniques. Let \( \gamma: \mathcal{T} \times \mathcal{R} \rightarrow \mathcal{T} \) denote the function for this operation:

\begin{align}
\gamma(X, \mathcal{R}) = \bigcup_{r \in \mathcal{R}} \text{extract\_pattern}(X, r)
\end{align}

Here, \( \text{extract\_pattern}(X, r) \) is a function that identifies and extracts patterns defined by the regular expressions \( r \) from the text \( X \).

\paragraph{Additional Text Cleaning Processes}
Further cleaning of the text involves removing redundant spaces and isolated characters. These processes are represented by the functions \( \delta: \mathcal{T} \rightarrow \mathcal{T} \) and \( \epsilon: \mathcal{T} \rightarrow \mathcal{T} \), respectively:

\begin{align}
\delta(X) &= \text{condense\_spaces}(X) \\
\epsilon(X) &= \text{filter\_chars}(X)
\end{align}

In these formulations, \( \text{condense\_spaces}(X) \) refers to the process of eliminating extra spaces, and \( \text{filter\_chars}(X) \) refers to the removal of isolated or irrelevant characters. These normalization and standardization steps are vital for preparing the text for subsequent analytical and modelling stages.

\subsection{Data Augmentation}

The data augmentation strategy for this project included collecting additional information through web scraping, an effective technique for expanding the dataset. This approach was motivated by the need to enrich the dataset with diverse examples to enhance the generalization of text classification models.

\paragraph{Web Scraping Collection and Process}
Web scraping was carried out on various retail websites, aligned with the nature of the study's data. Using Python libraries BeautifulSoup \cite{10145369}, we developed automated scripts to navigate product pages, extract relevant information such as name, description, category, and subcategory, and store them in a structured format. During the process, special attention was given to legal and ethical web scraping practices, respecting the websites' terms of use and avoiding server overload.

\paragraph{Integration and Harmonization of Collected Data}
The collected data was carefully integrated into the original dataset. A harmonization process was conducted to ensure consistency with the existing dataset's structure and format. This procedure involved normalizing categories and standardizing text formats to ensure data integrity and usability for subsequent analysis.

\paragraph{Impact on Diversity and Representativeness of the Dataset}
Adding this collected data significantly increased the diversity and representativeness of the dataset. The dataset became more balanced with a broader spectrum of product examples, vital for effectively training robust text classification models. Including web scraping data contributed to a better distribution of product classes.

This web scraping data augmentation approach proved valuable for enriching the dataset and improving the robustness and accuracy of the developed classification models.

\subsection{Model Selection and Configuration}
Two machine learning models were selected to address the task of multi-category text classification in the retail sector due to their specific capabilities in processing and understanding natural language: Long Short-Term Memory (LSTM) and BERT (Bidirectional Encoder Representations from Transformers).

\subsubsection{LSTM Model}
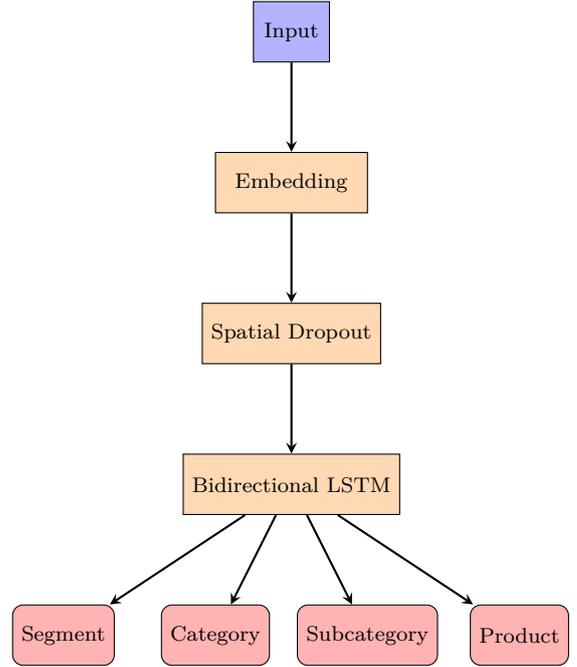
\begin{figure}[ht]
    \centering
    \begin{tikzpicture}[node distance=2cm]
        \node (input) [io] {Input};
        \node (embedding) [process, below of=input] {Embedding};
        \node (dropout) [process, below of=embedding] {Spatial Dropout};
        \node (lstm) [process, below of=dropout] {Bidirectional LSTM};

        \node (categoria) [startstop, below of=lstm, xshift=-3cm] {Segment};
        \node (segmento) [startstop, below of=lstm, xshift=-1cm] {Category};
        \node (subcategoria) [startstop, below of=lstm, xshift=1cm] {Subcategory};
        \node (produto) [startstop, below of=lstm, xshift=3cm] {Product};

        \draw [arrow] (input) -- (embedding);
        \draw [arrow] (embedding) -- (dropout);
        \draw [arrow] (dropout) -- (lstm);

        \draw [arrow] (lstm) -- (categoria);
        \draw [arrow] (lstm) -- (segmento);
        \draw [arrow] (lstm) -- (subcategoria);
        \draw [arrow] (lstm) -- (produto);
    \end{tikzpicture}
    \caption{LSTM network diagram for classification into Segment, Category, Subcategory, and Product.}
    \label{fig:lstm_network}
\end{figure}

The LSTM model was chosen for its ability to process data sequences and capture complex temporal dependencies, as illustrated in Figure \ref{fig:lstm_network}.

\paragraph{LSTM Architecture}
The LSTM architecture, represented in Figure \ref{fig:lstm_network}, consists of cells with three gates: input, forget, and output. The following mathematical equations describe these gates:

\begin{align}
i_t &= \sigma(W_{xi} x_t + W_{hi} h_{t-1} + W_{ci} c_{t-1} + b_i) \label{eq:input_gate} \\
f_t &= \sigma(W_{xf} x_t + W_{hf} h_{t-1} + W_{cf} c_{t-1} + b_f) \label{eq:forget_gate} \\
c_t &= f_t c_{t-1} + i_t \tanh(W_{xc} x_t + W_{hc} h_{t-1} + b_c) \label{eq:cell_state} \\
o_t &= \sigma(W_{xo} x_t + W_{ho} h_{t-1} + W_{co} c_t + b_o) \label{eq:output_gate} \\
h_t &= o_t \tanh(c_t) \label{eq:hidden_state}
\end{align}

These equations represent, respectively, the input gate (\ref{eq:input_gate}), forget gate (\ref{eq:forget_gate}), cell state (\ref{eq:cell_state}), output gate (\ref{eq:output_gate}), and hidden state (\ref{eq:hidden_state}) of an LSTM cell.

\paragraph{LSTM Training and Optimization}
The training of the LSTM model was executed with meticulous attention to detail, emphasizing the optimal selection of hyperparameters and the fine-tuning of algorithms to suit the specific nature of the data and the text classification task at hand. A vital component of this process was the incorporation of the Adam optimizer, an approach for stochastic optimization innovatively created by Kingma and Ba \cite{kingma2017adam}. The Adam optimizer is noted for its efficient handling of sparse gradients and adaptive estimation capabilities for first and second-order moments, making it exceptionally suitable for training sophisticated deep learning models like LSTM. Diverse configurations of the LSTM model and varied Adam optimizer settings were rigorously tested to identify the most effective combination for maximizing the model's performance in complex text classification tasks. Please see Appendix A for comprehensive details on the LSTM model's hyperparameters, architecture, and other configurations.

\paragraph{Brazilian Word Embeddings in LSTM}
To enhance the LSTM model's ability to understand and process the Portuguese language in retail contexts, we used Brazilian word embeddings from the NILC repository\footnote{http://nilc.icmc.usp.br/nilc/index.php/repositorio-de-word-embeddings-do-nilc}. Specifically, we employed Word2Vec CBOW with 50 dimensions and Glove with 50, 100, and 1000 dimensions. These pre-trained embeddings, adapted to Portuguese, provided a solid foundation for the model to capture linguistic and semantic nuances essential for text classification \cite{hartmann2017portuguese}.

\paragraph{LSTM Text Classification Implementation}
The LSTM model for text classification was implemented using TensorFlow and Keras, leveraging the capabilities of these libraries for efficient neural network modelling and training. The LSTM architecture was adapted to handle the specific characteristics of retail data, ensuring that the model could accurately process and classify texts.

\subsubsection{BERT Model}
The BERT model was selected due to its exceptional ability to understand the bidirectional context in text, making it ideal for short-text classification.

\begin{figure}[ht]
    \centering
    \begin{tikzpicture}[node distance=2cm]
        \node (input) [io] {Input};
        \node (bert) [process, below of=input] {BERT Layers};

        \node (categoria) [startstop, below of=bert, xshift=-3cm] {Segment};
        \node (segmento) [startstop, below of=bert, xshift=-1cm] {Category};
        \node (subcategoria) [startstop, below of=bert, xshift=1cm] {Subcategory};
        \node (produto) [startstop, below of=bert, xshift=3cm] {Product};

        \draw [arrow] (input) -- (bert);
        
        \draw [arrow] (bert) -- (categoria);
        \draw [arrow] (bert) -- (segmento);
        \draw [arrow] (bert) -- (subcategoria);
        \draw [arrow] (bert) -- (produto);
    \end{tikzpicture}
    \caption{BERT Model Diagram with Identical Input and Output.}
    \label{fig:bert_model}
\end{figure}
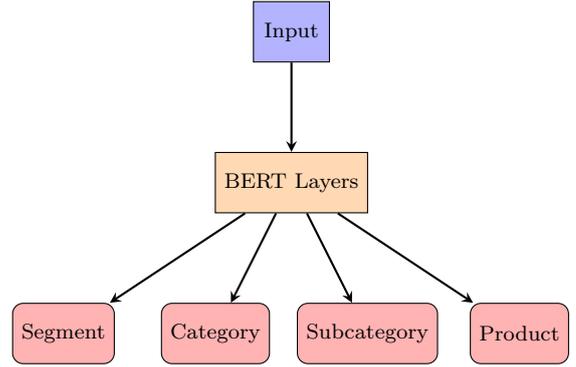

\paragraph{BERT Architecture}
The BERT architecture depicted in Figure \ref{fig:bert_model} is based on the innovative Transformer structure, which uses attention mechanisms to capture contextual relationships between words in a text. The central element of this architecture is the attention mechanism, described by the following equation:

\begin{equation}
\text{Attention}(Q, K, V) = \text{softmax}\left(\frac{QK^T}{\sqrt{d_k}}\right)V \label{eq:bert_attention}
\end{equation}
where \(Q\), \(K\), and \(V\) represent the Query, Key, and Value matrices, respectively, and \(d_k\) is the dimension of the keys. Equation \ref{eq:bert_attention} describes the fundamental calculation within the BERT attention mechanism.

\paragraph{BERT Training and Optimization}
The training of our BERT model utilized BERTimbau Base, a pre-trained version of BERT specifically tailored for Brazilian Portuguese. Developed by Souza, Nogueira, and Lotufo, BERTimbau Base provides an optimized foundation for understanding and processing Brazilian Portuguese texts, making it distinctly suited for our study focused on the retail sector in Brazil \cite{souza2020bertimbau}. The model training involved advanced techniques, including the Masked Language Model (MLM) and Next Sentence Prediction (NSP). Model optimization was further enhanced through careful hyperparameter tuning, leveraging the advanced computational capabilities and optimization techniques inherent to transformer models. For comprehensive details on the BERT model's hyperparameters, architecture, and other configurations, please see Appendix B.

\paragraph{BERT Text Classification Implementation}
For text classification in the retail sector, BERTimbau Base was implemented using the Transformers library from Hugging Face\footnote{https://huggingface.co/neuralmind/bert-base-portuguese-cased}, a popular framework for working with transformer-based models. This choice was due to its flexibility and efficiency in handling such models. The implementation was specifically adapted to meet the unique requirements of our task, focusing on predicting multiple categories within the retail sector. The performance of the BERTimbau Base model was meticulously evaluated using standard NLP metrics to ensure its effectiveness in accurately classifying Brazilian Portuguese retail product descriptions.

\subsection{Metrics and Loss}
The F1 Macro Score metric was employed to evaluate classification models, in addition to using focal loss as the loss function. The F1 Macro Score notably applies to class imbalances, as it averages the balance between precision and recall across all classes.

\subsubsection{F1 Macro Score Metric}
The F1 Macro Score \cite{inproceedings} adapts the traditional F1 Score, tailored for multi-class imbalance scenarios. It calculates the F1 Score separately for each class and then averages them, giving equal weight to each class. This is remarkably beneficial in datasets where some classes are underrepresented. Mathematically, the F1 Macro Score is expressed as:

\begin{equation}\label{eq:f1_macro_score}
    F1_{\text{macro}} = \frac{1}{N} \sum_{i=1}^{N} 2 \times \frac{\text{precision}_i \times \text{recall}_i}{\text{precision}_i + \text{recall}_i}
\end{equation}

In this equation, \( N \) represents the number of classes, and \( \text{precision}_i \) and \( \text{recall}_i \) are the precision and recall calculated for each class \( i \). The F1 Macro Score thus provides a more comprehensive evaluation of the model's performance across all classes.

\subsubsection{Focal Loss}
Cross-entropy \cite{10.5555/3618408.3619400} is often used as a loss function in classification problems, but it can be inadequate in cases of severe class imbalance. Focal loss \cite{lin2017focal} was introduced to mitigate this problem by adjusting the contribution of each class to the total loss based on how easily the model classifies the samples. Focal loss is mathematically expressed as:

\begin{equation}\label{eq:focal_loss}
    FL(p_{t}) = - (1 - p_{t})^{\gamma} \log(p_{t})
\end{equation}

where \( p_{t} \) is the predicted probability for the true class, and \( \gamma \) is the focal parameter. This approach reduces the weight of classes with high incidence and increases the weight of underrepresented classes.

\begin{figure}[!hbt]
    \centering
    \includegraphics[width=\columnwidth]{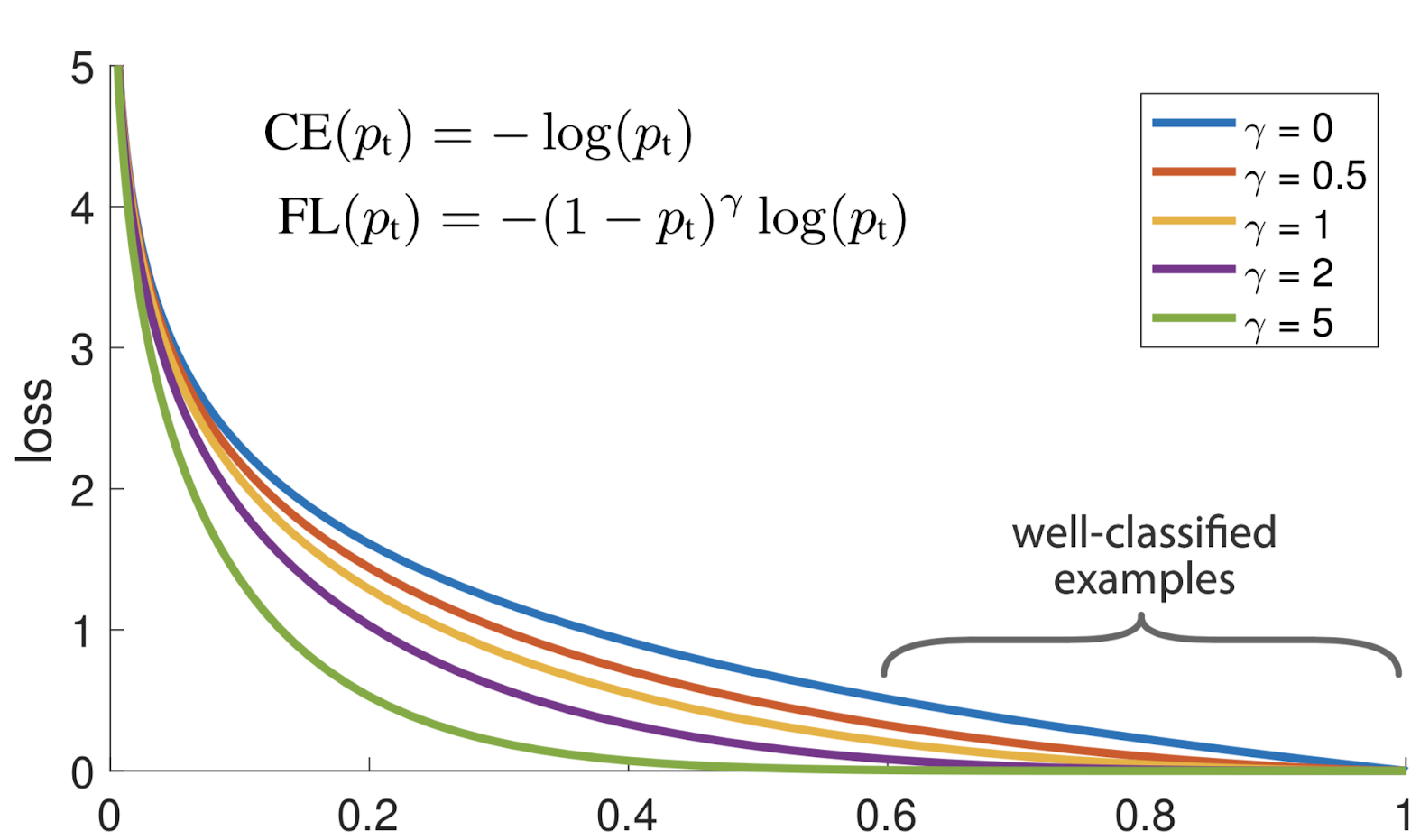}
    \label{fig:focal_loss}
    \caption{Behavior of Focal Loss for Different \( \gamma \) Values}
\end{figure}

\subsection{Final Considerations on the Methodology}
This section concludes with a discussion of the methodological choices made during the research. It will address the reasons for selecting specific techniques and models, the limitations encountered and how they were mitigated. Using the F1 Score metric, in conjunction with focal loss, provides an effective means to evaluate and optimize classification models,  exceptionally in scenarios with class imbalance, as is often the case in retail product classification. The adopted methodology aims to ensure that the results are technically valid, applicable, and relevant to the retail sector.

\section{Results and Discussion}

This section presents and discusses the results of applying LSTM and BERT models for text classification focused on predicting multiple categories within the retail sector. We used test data sets to analyze these models' effectiveness rigorously. The results regarding model performance are examined, focusing on the impact of data augmentation techniques and focal loss. These findings are compared against the initial study's expectations and objectives, providing a comprehensive understanding of the model's capabilities in practical scenarios.

First, we explore the results obtained with the LSTM model, including a detailed analysis of the model's performance in the F1-score metric. Next, we present the results of the BERT model, focusing on how its architecture and specific features influenced text classification.

Furthermore, this section discusses the impact of focal loss and data augmentation techniques on model performance. These analyses are essential to understanding the effectiveness of the adopted approaches and identifying areas for improvement and optimization for future research.

The results are then contextualized within the NLP and text classification field, highlighting their contributions to the area. This discussion aims to validate the results and explore their practical and theoretical implications in the broader context of retail and natural language processing technology.

\subsection{Analysis of LSTM Model Results}

The results obtained with the LSTM model, employing different Word Embedding strategies, are presented in Table \ref{tab:resultados_lstm}. These results are essential to understanding the impact of different word representations on text classification performance to predict multiple categories in the retail sector.

\begin{table}[ht]
\centering
{\fontfamily{phv}\selectfont} % Initiates Helvetica font
\tiny
\caption{LSTM Model Results with Different Word Embedding Strategies using Cross-entropy Loss}
\begin{tabular}{l| c c c c}
\hline
\textbf{Word Embedding} & \textbf{Segment} & \textbf{Category} & \textbf{Subcategory} & \textbf{Product} \\ \hline
Word2Vec s100 & 89.2 & 63.7 & 57.3 & - \\
GloVe s50 & 93.5 & 70.8 & 68.6 & 64.3 \\ 
\textbf{GloVe s100} & \textbf{93.8} & \textbf{72.5} & \textbf{69.1} & \textbf{65.1} \\
GloVe s1000 & 93.8 & 70.4 & 69.3 & 64.7 \\ \hline
\end{tabular}
\label{tab:resultados_lstm}
\end{table}

A detailed analysis of the results reveals that the model with pre-training using GloVe s100 achieved the highest F1 score in all categories, except for the subcategory, where GloVe s1000 performed slightly better. This suggests that the dimensionality of the embedding plays a significant role in the model's effectiveness, with a higher number of dimensions providing a richer and more nuanced representation of words.

On the other hand, Word2Vec (CBOW s100) proved to be less effective, especially in segment and category classification. This could be due to inherent differences in the Word2Vec and GloVe techniques in capturing semantic contexts and word relationships.

It is important to note that the absence of data for the 'Product' category using Word2Vec (CBOW s100) was because we did not initially use this category.

These results highlight the importance of choosing an appropriate Word Embedding technique in developing NLP models, especially in practical applications such as text classification to predict multiple categories in the retail sector.

\subsection{Analysis of BERT Model Results}

Applying the BERT model to the text classification task to predict multiple categories in the retail sector produced remarkable results. The model achieved an F1-score of 91.2\% for 'Segment', 79.3\% for 'Category', 79.1\% for 'Subcategory', and 78.2\% for 'Product'. These results are a testament to the effectiveness of BERT's bidirectional attention architecture in understanding texts' complex context and nuances.

\begin{table}[ht]
\centering
{\fontfamily{phv}\selectfont} % Initiates Helvetica font
\tiny
\caption{BERT Model Results and Best LSTM Model using Cross-entropy Loss}
\begin{tabular}{l | c c c c}
\hline
\textbf{Model} & \textbf{Segment} & \textbf{Category} & \textbf{Subcategory} & \textbf{Product} \\ \hline
LSTM & 93.8 & 72.5 & 69.1 & 65.1 \\ 
\textbf{BERT} & \textbf{91.2} & \textbf{79.3} & \textbf{79.1} & \textbf{78.2} \\ \hline
\end{tabular}
\label{tab:resultados_bert}
\end{table}

Compared to the results obtained by the LSTM model, BERT showed superior performance in more specific categories, such as 'Product'. This indicates that its ability to understand detailed contexts is highly beneficial for categories that require a more granular analysis. Alternatively, LSTM performed comparably in higher-level categories, such as 'Segment', suggesting that different models may be more efficient depending on the nature of the classification task.

These findings reinforce the importance of selecting the most suitable natural language processing model for each specific application, taking into account the data's particular characteristics and the classification task's requirements.

\subsection{Impact of Focal Loss and Data Augmentation}

The application of focal loss and data augmentation, including data collection via web scraping from other retail websites, significantly improved the LSTM and BERT classification models.

\begin{table}[ht]
\centering
\tiny
\caption{Improved Model Results with Data Augmentation and Focal Loss}
\begin{tabular}{l| c c c c}
\hline
\textbf{Model} & \textbf{Segment} & \textbf{Category} & \textbf{Subcategory} & \textbf{Product} \\ \hline
LSTM Aug & 95.7 & 90.4 & 91.8 & 86.4 \\ 
LSTM FL and Aug & 97.2 & 92.6 & 92.7 & 88.8 \\ 
BERT Aug & 98.6 & 94.9 & 95.2 & 91.5 \\ 
\textbf{BERT FL and Aug} & \textbf{99.1} & \textbf{96.3} & \textbf{96.4} & \textbf{93.0} \\ \hline
\end{tabular}
\label{tab:resultados_melhorados}
\end{table}

The collection of additional data significantly increased the original dataset, adding approximately 30,000 records. This enlargement of the dataset expanded the amount of information available for training but also increased the diversity and representativeness of product categories, improving the model's accuracy.

The introduction of focal loss had a particularly positive impact on the BERT model, as evidenced by the results in Table \ref{tab:resultados_melhorados}. Focal loss helped mitigate the effect of class imbalance, allowing the model to focus more effectively on less represented classes. This significantly boosted F1 scores in all categories, notably in more detailed ones like 'Product.'

These improvements spotlight the value of combining advanced modelling techniques with data preprocessing and augmentation strategies, demonstrating how these approaches can effectively overcome common challenges in NLP tasks, such as class imbalance and limitations in data availability.

\subsection{General Discussion and Implications of Results}

The results obtained in the LSTM and BERT models, especially after incorporating focal loss and data augmentation techniques, offer valuable comprehension of the effectiveness of different NLP modelling and preprocessing approaches. These results have significant implications for short text classification in the retail sector, highlighting the importance of choosing modelling and preprocessing techniques appropriately based on the specific characteristics of the data and tasks.

Comparing our findings with existing literature, we note alignment with current trends in NLP, emphasizing the importance of advanced models and sophisticated preprocessing techniques. The effectiveness of BERT models in various NLP tasks is well-documented, and our results reinforce its applicability in specific contexts such as retail. Likewise, focal loss and data augmentation have proven efficient methods for addressing challenges such as class imbalance and data limitations.

These results also raise important considerations about the selection and application of NLP models. For example, the choice between LSTM and BERT should consider the nature of the classification task and dataset characteristics. Similarly, the use of focal loss and data augmentation should be weighed against the specific needs of each project.

This discussion points to the need for a careful and contextualized approach in developing NLP solutions, suggesting paths to optimize models and techniques according to the specific requirements of each application.

\subsection{Limitations and Suggestions}

Despite the promising results obtained in this study, some limitations should be acknowledged. One of the main limitations is the dependence on the quality and variety of the collected data, which can influence model generalization. Additionally, the effectiveness of focal loss and data augmentation in other NLP contexts still requires further investigation.

We suggest exploring more diverse and extensive datasets for future research to improve model generalization. Investigating the applicability of focal loss and data augmentation techniques in different NLP domains would also be beneficial, especially in scenarios with more extreme class imbalances. Another area of interest would be experimentation with newer and advanced NLP models and techniques.

\subsection{Conclusion of the Results and Discussion}

This section provided a detailed analysis of the results obtained with the LSTM and BERT models in the context of text classification to predict multiple categories in the retail sector. Implementing focal loss and data augmentation techniques proved effective, significantly improving the models' performance, especially in class imbalance situations.

The encouraging results highlight the importance of advanced modelling and data preprocessing techniques in NLP. This study contributes to natural language processing by demonstrating the applicability and effectiveness of complex models and preprocessing techniques in a practical and challenging context.

In conclusion, the results obtained in this study provide valuable insights for the NLP community and establish a solid foundation for future research in text classification in specific sectors such as retail.

\section{Conclusion and Future Work}

This study presented a detailed analysis of using LSTM and BERT models for text classification to predict multiple categories in the retail sector, demonstrating the effectiveness of advanced modelling and data preprocessing techniques. Key innovations included the application of focal loss and data augmentation, which was demonstrated to be crucial for improving model performance in class imbalance scenarios. The results revealed significant improvements in classification accuracy, highlighting the potential of these techniques in practical NLP applications.

The contributions of this study are particularly relevant to the field of natural language processing, offering insights into how complex models and preprocessing strategies can be effectively applied to overcome common challenges in text analysis. Additionally, the study provides a solid foundation for applying NLP in the retail sector, a domain with specific demands and unique challenges.

For future work, exploring even larger and more diverse datasets, including data from different sources and languages, is suggested to further enhance model generalization and effectiveness. Furthermore, it would be valuable to investigate the applicability of new modelling and preprocessing techniques, especially those emerging in the rapidly evolving field of NLP. Lastly, studying specific strategies to handle extreme class imbalances in other application domains would also be a fruitful area for future research.

This study represents an essential step in understanding the application of NLP in the retail sector, paving the way for future investigations and developments in the field.

% Encerramento do documento com agradecimentos e referências, se aplicável.
\section*{Acknowledgments}
Daniel C. Fuentes Guzman acknowledges the support of the Coordination for the Improvement of Higher Education Personnel - Brazil (CAPES) - Financial Code 001. Francisco Louzada is supported by the Brazilian agencies CNPq (grant number 301976/2017-1) and FAPESP (grant number 2013/07375-0).

\appendix
\section{LSTM Model Configurations}

This appendix details the configurations and architecture of the LSTM model used in the study, including applying various word embedding strategies and implementing the Focal Loss function.

\subsection{Model Hyperparameters}
\begin{table}[ht]
\scriptsize
\centering
\caption{Hyperparameters of the LSTM Model}
\begin{tabular}{|l|l|}
\hline
\textbf{Parameter} & \textbf{Value} \\
\hline
Embedding Dimension & 100 \\
Maximum Sequence Length & 38 \\
Maximum Number of Words & 42000 \\
Optimizer & Adam \\
Learning Rate & 1e-5 \\
Batch Size & 64 \\
Epochs & 50 \\
Early Stop & 3 \\
\hline
\end{tabular}
\label{tab:hyperparameters_lstm}
\end{table}

\subsection{Neural Network Architecture}
The architecture of the LSTM model was structured as follows:

\begin{table}[ht]
\centering
\caption{Architecture of the LSTM Model}
\scriptsize
\begin{tabular}{|l|l|}
\hline
\textbf{Layer Type} & \textbf{Configuration} \\
\hline
Embedding Layer & Trainable Weights \\
Spatial Dropout & 0.2 \\
Bidirectional LSTM 1 & 100 units, dropout=0.2\\
Bidirectional LSTM 2 & 200 units, dropout=0.2 \\
\hline
\end{tabular}
\label{tab:model_architecture}
\end{table}

This table provides a detailed view of each layer in the LSTM model, including the type and specific configurations for each part of the network.

\subsection{Word Embeddings Used}
Different word embeddings were tested to evaluate the model's performance. The configurations were:

\begin{table}[ht]
\centering
\caption{Word Embeddings Tested in the LSTM Model}
\scriptsize
\begin{tabular}{|l|c|}
\hline
\textbf{Word Embedding} & \textbf{Dimension} \\
\hline
Word2Vec (CBOW s100) & 100 \\
GloVe s50 & 50 \\
GloVe s100 & 100 \\
GloVe s1000 & 1000 \\
\hline
\end{tabular}
\label{tab:word_embeddings_lstm}
\end{table}

\subsection{Focal Loss Configuration}
\begin{table}[ht]
\centering
\caption{Focal Loss Parameters for LSTM Model}
\scriptsize
\begin{tabular}{|l|l|}
\hline
\textbf{Parameter} & \textbf{Value} \\
\hline
From Logits & False \\
Alpha & 0.25 \\
Gamma & 2.0 \\
\hline
\end{tabular}
\label{tab:focal_loss_parameters}
\end{table}

This table outlines the specific parameters used in the Sigmoid Focal Cross-Entropy function for optimizing the LSTM model. The configuration was chosen to address the class imbalance effectively and enhance the model's learning from under-represented data.

This configuration of the Focal Loss function was critical in addressing class imbalance, allowing the model to focus more effectively on challenging classes and enhancing its performance on under-represented data.

\section{BERT Model Configurations}

This appendix details the configurations and architecture of the BERT model used in the study, including the application of various hyperparameters and the implementation of the Focal Loss function.

\subsection{Model Hyperparameters}
\begin{table}[ht]
\scriptsize
\centering
\caption{Hyperparameters of the BERT Model}
\begin{tabular}{|l|l|}
\hline
\textbf{Parameter} & \textbf{Value} \\
\hline
Pre-trained Model & BERTimbau \\
Maximum Sequence Length & 38 \\
Optimizer & AdamW \\
Learning Rate & 5e-5 \\
Batch Size & 32 \\
Epochs & 40 \\
Early Stop & 10 \\
\hline
\end{tabular}
\label{tab:hyperparameters_bert}
\end{table}

\subsection{Neural Network Architecture}
The BERT model used in this study is the BERTimbau Base, specifically the 'neuralmind/bert-base-portuguese-cased' model. This model is a pre-trained BERT model for Brazilian Portuguese and achieves state-of-the-art performance in various NLP tasks. The specifics of the model used are as follows:

\begin{table}[ht]
\centering
\caption{Architecture Details of BERTimbau Base Model}
\scriptsize
\begin{tabular}{|l|l|}
\hline
\textbf{Aspect} & \textbf{Configuration} \\
\hline
Model & bert-base-portuguese-cased \\
Architecture Type & BERT-Base \\
Number of Layers & 12 \\
Number of Parameters & 110M \\
\hline
\end{tabular}
\label{tab:model_architecture_bertimbau}
\end{table}

The BERT-Base architecture consists of 12 transformer layers, with each layer having a hidden size of 768 and 12 self-attention heads. The model has a total of 110 million parameters, making it suitable for a wide range of NLP tasks, including Named Entity Recognition, Sentence Textual Similarity, and Recognizing Textual Entailment.

\subsection{Focal Loss Configuration}
The notebook includes the implementation of Focal Loss for the BERT model. The specific parameters used for Focal Loss are as follows:

\begin{table}[ht]
\centering
\caption{Focal Loss Parameters for BERT Model}
\scriptsize
\begin{tabular}{|l|l|}
\hline
\textbf{Parameter} & \textbf{Value} \\
\hline
Gamma1 & 2 \\
Gamma2 & 1 \\
Gamma3 & 1 \\
Gamma4 & 2 \\
Dropout & 0.5 \\
\hline
\end{tabular}
\label{tab:focal_loss_parameters_bert}
\end{table}

% Bibliografia
\bibliography{reference}

\end{document}